\documentclass[runningheads]{llncs}
\usepackage{graphicx}
\usepackage{multirow}
\usepackage[caption=false,font=footnotesize]{subfig}
\usepackage[dvipsnames]{xcolor}

\begin{document}
\title{Hybridised Loss Functions for Improved Neural Network Generalisation\thanks{This research was funded by the National Research Foundation of South Africa (Grant Number: 120837).}}
\author{Matthew C. Dickson\inst{1} \and
Anna S. Bosman\inst{1}\orcidID{0000-0003-3546-1467} \and
Katherine M. Malan\inst{2}\orcidID{0000-0002-6070-2632}}
\authorrunning{M. Dixon et al.}
\institute{Department of Computer Science, University of Pretoria, Pretoria, South Africa  
\email{anna.bosman@up.ac.za}
\and
Department of Decision Sciences, University of South Africa, Pretoria, South Africa\\
\email{malankm@unisa.ac.za}}
\maketitle              % typeset the header of the contribution
\begin{abstract}
Loss functions play an important role in the training  of  artificial neural networks (ANNs), and can affect the generalisation ability of the ANN model, among other properties. Specifically, it has been shown that the cross entropy and sum squared error loss functions result in different training dynamics, and exhibit different properties that are complementary to one another. It has previously been suggested that a hybrid of the entropy and sum squared error loss functions could combine the advantages of the two functions, while limiting their disadvantages. The effectiveness of such hybrid loss functions is investigated in this study. It is shown that hybridisation of the two loss functions improves the generalisation ability of the ANNs on all problems considered. The hybrid loss function that starts training with the sum squared error loss function and later switches to the cross entropy error loss function is shown to either perform the best on average, or to not be significantly different than the best loss function tested for all problems considered. This study shows that the minima discovered by the sum squared error loss function can be further exploited by switching to cross entropy error loss function. It can thus be concluded that hybridisation of the two loss functions could lead to better performance in ANNs

\end{abstract}
\section{Introduction}
Loss functions in artificial neural networks (ANNs) are used to quantify the error produced by the model on a given dataset. ANNs are trained via the minimisation of a given loss function. Therefore, loss function properties can directly affect the properties of the resulting ANN model~\cite{2,3}. One such property that is of a specific interest is the ability of the ANN to generalise, i.e. correctly predict the outputs for data patterns not seen during training. It was previously shown that choosing an appropriate loss function for a problem is a fundamental task in ANN design, as it has a direct correlation to model performance during evaluation~\cite{2}.

Techniques that deal with training an ANN in order to develop better performing models usually involve tuning hyper-parameters and selecting an architecture that best suits the problem being solved in any given instance~\cite{8}. However, another factor that could potentially improve the performance of ANNs is choosing the correct loss function for a problem, as there is no single universal loss function that performs the best on all problems~\cite{2}. It is therefore suggested that adapting existing loss functions that have complementary properties and combining them to form a new hybrid error metric may improve or mitigate faults of existing loss functions, and thus create adaptable general loss functions~\cite{2,3}. These hybrid loss functions can then be used on a range of problems with varying degrees of difficulty and ultimately produce better ANN models.

This study investigates a number of different hybrid loss functions that combine the entropic and the quadratic loss functions. Hybrid variants investigated include static combinations of the two loss functions with different proportions, combinations that gradually adapt from one function to the other, and hybrids that switch from one function to the other based on stagnation/deterioration. The proposed hybrids are tested on a selection of classification problems and benchmarked against the constituent entropic and quadratic loss functions. Results show that the hybrid loss functions generalised better or on a par with the baseline loss functions on all problems considered.  It is also shown that the hybrid
%loss function variant which began training using the quadratic loss, and later switched over to the entropic loss performed better or on par with the non-hybrid loss functions.
which starts training using the quadratic loss, and later switches over to the entropic loss, performed the best overall. Contrary to the results of Golik \emph{et al.} \cite{3}, who found that the quadratic loss did not perform well with randomised weights, but performed well when used to fine-tune the solution that the entropic loss function produced, this study shows that the minima discovered by quadratic loss can be further exploited by switching to entropic loss.

The rest of the paper is organised as follows. Section~\ref{sec:loss} defines the loss functions hybridised in this study. Section~\ref{sec:method} proposes three types of hybrid losses. Section~\ref{sec:setup} details the experimental setup. Section~\ref{sec:results} provides the empirical results, and Section~\ref{sec:discuss} discusses them. Finally, Section~\ref{sec:conclude} concludes the paper.

\section{Loss Functions in ANNs}\label{sec:loss}
In order for an ANN to solve a particular problem, a loss function must be defined. Loss functions are used to measure the error of a model. Two commonly used error metrics for determining the effectiveness of a classification model are the sum squared error (SE), also known as quadratic error, and the cross entropy error (CE) loss functions~\cite{2,22}. SE and CE are formally defined below in Sections~\ref{subsec:se} and~\ref{subsec:ce}, respectively. Section~\ref{subsec:effect} discusses the effect that the loss functions can have on ANN performance.

\subsection{Sum Squared Error}\label{subsec:se}
The SE function  is defined as:
\begin{equation}
E_{se}=\sum_{p=1}^{P}\sum_{k=1}^{K}(t_{k,p}-o_{k,p})^2.
\label{eq1}
\end{equation}
For Equations \ref{eq1} and \ref{eq2}, $P$ is the total number of training patterns, $K$ is the total number of output units, $t_{k,p}$ is the $k$-$th$ target value for pattern $p$, and $o_{k,p}$ is the $k$-$th$ output obtained for pattern $p$. The SE function, also known as the quadratic loss, calculates the sum of the squared errors produced by an ANN during training. The minimisation of the SE lowers the total error produced by an ANN.
\subsection{Cross Entropy}\label{subsec:ce}
The CE function is defined as:
\begin{equation}
E_{ce}=\sum_{p=1}^{P}\sum_{k=1}^{K}\big(t_{k,p}\log o_{k,p}+(t_{k,p}-1) \log (o_{k,p}-1)\big).
\label{eq2}
\end{equation}
The CE function, also known as the entropic loss, measures the difference between two distributions, namely the distribution of the target outputs of the ANN and the distribution of the actual outputs of the observations in the dataset. The entropic loss can only be used if the outputs of an ANN can be interpreted as probabilities. The minimisation of the CE results in the convergence of the two distributions.

\subsection{The Effect of Loss Functions on Performance}\label{subsec:effect}
Bosman \emph{et al.} \cite{2} have shown that a loss function directly impacts the performance of an ANN  for a given problem, and therefore is a major factor to consider when implementing an ANN model. It has been observed  that the true posterior probability for both the CE and SE loss functions is a global minimum and as such, in theory an ANN can be equally trained by minimising either of the two criteria as long as it is able to approximate the true posterior distribution within an arbitrary close range \cite{9,3}. However, in practice this is not the case, as the SE and CE loss functions exhibit different properties from one another, leading to different results in accuracy values during both training and testing \cite{3}. 

It has also been found that the SE loss function is more resilient to overfitting than the CE loss function, due to the fact that the SE loss function is bounded when modelling a distribution, therefore minimisation is more robust to outliers compared to the minimisation of the CE loss function \cite{3}. Resilience to overfitting indicates that SE may exhibit superior generalisation properties compared to CE. However, the entropic loss exhibits stronger gradients than the quadratic loss, therefore convergence is faster for the entropic loss~\cite{2}. It has also been argued \cite{2,3} that the entropic loss provides a more searchable loss landscape due to larger gradients, as compared to the quadratic loss, and thus should be more favourable to gradient-based optimisation techniques such as stochastic gradient descent.

In terms of the generalisation behaviour, it has been suggested in the literature that wide minima generalises better than narrow minima~\cite{chaudhari2019entropy,keskar2016large}. Bosman \emph{et al.} \cite{2} showed that SE exhibits more local minima than CE, thus making CE less susceptible to local minima trapping. However, the stronger gradients of CE suggest that CE is more likely to converge to sharp (narrow) minima with inferior generalisation performance.

\section{Hybrid Loss Functions}\label{sec:method}
A number of hybrid variants of the entropic and quadratic loss functions are proposed in this section to investigate whether such hybrids could combine the advantages while limiting the disadvantages of the two loss functions. With the hybrid approach, a new error metric can be developed and used for the purpose of training and evaluating ANNs. This hybrid metric may result in better generalisation and improved loss landscape properties, leading to loss functions that yield better ANN models.

It was observed \cite{3} that, if started with good initial ANN weights, the SE loss function could on average further improve the minimisation of the error value found by the CE loss function in different system setups. This observation once again indicates that a hybrid approach to loss functions in ANN training could have merit.

This study provides an empirical investigation into different ways of combining the SE and CE loss functions. Nine different hybrid variants of loss functions were tested, summarised in Table \ref{tab:lossfunctions}. Variants 1 and 5 are the basic CE and SE loss functions, respectively, which were included as baselines for comparison against the different hybrids. The remaining hybrids are described below.

For the purpose of this study, three different approaches to hybridise the two loss functions were experimented with, which will be described as static (Section~\ref{subsec:static}), adaptive (Section~\ref{subsec:adaptive}), and reactive (Section~\ref{subsec:reactive}). 
%We experimented with three different approaches on how to hybridize the two loss functions which we will describe as static, adaptive and reactive. %

\subsection{Static approach}\label{subsec:static}
In the static approach, the loss is defined as a weighted sum of the quadratic (SE) and entropic (CE) loss functions. The error values from Equations \ref{eq1} and \ref{eq2} are not normalised and the range of values are problem dependent. Therefore, to avoid an imbalance in the sum, the error values from the individual loss functions are first normalised using an estimate of the maximum value for the loss on that problem.

The hybrid loss function used is defined as follows:
\begin{equation}
E_{he} = s_{1}\frac{E_{se}}{max_{se}}+s_{2}\frac{E_{ce}}{max_{ce}}
\label{eq3}
\end{equation}
where $E_{se}$ is the sum square error loss function, $E_{ce}$ is cross entropy loss function as defined in Equations (\ref{eq1}) and (\ref{eq2}) respectively, $s_1$ and $s_2$ are scalar values that are the proportions given to the loss functions where  $s_1 + s_2 = 1$. The values $max_{se}$ and $max_{ce}$ are the approximate maximum values produced by the individual loss functions over the problem before training.

%The proportions that were used for this analysis are defined in Table~\ref{tab:lossfunctions}, where each static hybrid is assigned a unique label.
The proportions that were used for this analysis are defined in Table~\ref{tab:lossfunctions}, where each static hybrid is assigned a unique label. The proportions for the static hybrid loss functions were chosen based on linearly subdividing the range, producing three static hybrid loss functions that could be tested.

\begin{table}[!t]
	\caption{Nine variants of the loss functions used in the experimentation}
	\label{tab:lossfunctions}
\centering
	\begin{tabular}{ |l | l |p{7.5cm}|}
		\hline
		
		Variant & Label & Description \\ \hline
		1. CE only & $CE_{100}SE_{0}$ & Equation \ref{eq3} with $s_1=0$, $s_2=1$\\
		2. Static hybrid 1 & $CE_{75}SE_{25}$ & Equation \ref{eq3} with $s_1 = 0.25$, $s_2 = 0.75$ \\
		3. Static hybrid 2 & $CE_{50}SE_{50}$ & Equation \ref{eq3} with $s_1 = 0.5$, $s_2 = 0.5$\\ 
	    4. Static hybrid 3 & $CE_{25}SE_{75}$ & Equation \ref{eq3} with $s_1 = 0.75$, $s_2 = 0.25$\\ 
	    5. SE only & $CE_{0}SE_{100}$ & Equation \ref{eq3} with $s_1 = 1$, $s_2 = 0$\\ 
		6. Adaptive hybrid 1 & $CE_{to}SE$ & Starting with 100\% CE and moving over to 100\% SE in steps of 1\%\\ 
		7. Adaptive hybrid 2 & $SE_{to}CE$ & Starting with 100\% SE and moving over to 100\% CE in steps of 1\%\\
		8. Reactive hybrid 1 & $CE_{>>}SE$ & Starting with CE and switching to SE on stagnation / deterioration \\
		9. Reactive hybrid 2 & $SE_{>>}CE$ & Starting with SE and switching to CE on stagnation / deterioration  \\\hline
	\end{tabular}
\end{table}

\subsection{Adaptive approach}\label{subsec:adaptive} In the adaptive approach, the same loss function as defined in Equation~(\ref{eq3}) was used, but instead of assigning a static proportion to each component of the hybrid at the beginning of training, either SE or CE was chosen to contribute 100\% to the loss function at the start (either $s_1$ or $s_2$ set to 1). The proportion of that component was gradually decreased by a factor of one percent for each epoch until that component had no weight associated with it, while the proportion of the other component was increased by the same factor until the last epoch. 

In the experiments, training was executed for one hundred epochs, so the adaptation was spread over the full length of the training. This approach produced two hybrids, $CE_{to}SE$ and $SE_{to}CE$, where the $CE_{to}SE$ hybrid gradually changed from CE to SE, and the $SE_{to}CE$ hybrid gradually changed from SE to CE.

%For the Reactive approach: For this approach we started with only MSE or CE then after 20 epochs of training we checked to see if any stagnation or deterioration occurred in the training. If after 3 epochs such observations were seen then we switched over to MSE or CE depending on which loss function we started with and then continued training until 100 epochs.%

\subsection{Reactive approach}\label{subsec:reactive} 
Two reactive hybrids are proposed that switch from one loss function to the other based on the training error. At the beginning of training, only SE or CE was used for the training of the ANN, then after twenty epochs of training a condition was checked to see if any stagnation or deterioration occurred in the accuracy of the ANN. If no improvement occurred after three sequential epochs, then a switch was performed from one baseline function to the other depending on which loss function was deployed at the beginning of training. This approach yielded two hybrid loss functions, $CE_{>>}SE$ and $SE_{>>}CE$, where the hybrid $CE_{>>}SE$ loss function began training with CE, and the hybrid $SE_{>>}CE$ loss function began training with SE.

\section{Experimental Setup}\label{sec:setup}
This section details the experimental setup of the study. Section~\ref{subsec:data} lists the datasets used in the experiments, and Section~\ref{subsec:arch} discusses the hyper-parameters chosen for the ANNs.

\subsection{Datasets}\label{subsec:data}
The following five classification datasets were used in this study:
\begin{enumerate}
	\item \textbf{Cancer:} Consists of 699 observations each containing a tumor. Depending on the values of 30 features, the observations are classified into tumors either being benign or malignant~\cite{15}.
	\item \textbf{Glass:} Consists of 214 observations of glass shards. Each glass shard belongs to one of six classes, depending on nine features that capture the chemical components of the glass shards. The classes include float processed or non-float process building windows, vehicle windows, containers, table-ware, or head lamps~\cite{15}.   
	\item \textbf{Diabetes:} Consists of 768 observation of Pima Indian patients. Depending on the values of eight features, observations are classified as either being diabetic or not~\cite{15}. 
	\item \textbf{MNIST:} Consists of 70 000 observations of handwritten digits, where each digit is a 28 by 28 pixel image of a digit between 0 and 9 \cite{16}.
	\item \textbf{Fashion-MNIST:} Consists of 70 000 observations that include ten categories of clothing, where each item is a 28 by 28 gray-scale pixel image  \cite{25}. 
\end{enumerate}

%\begin{description}
%	\item[1.] MNIST : Contains 70000 observations that are handwritten digits each is a 28 by 28 pixel image of a digit between 0 and 9 \cite{16}.
%	\item[2.] Diabetes: Contains 768 observation of Pima indian patients. Depending on the values of 8 features are then classified into either being diabetic  or not \cite{15}.   
%	\item[3.] Fashion-MNIST : Contains 70000 observations that includes ten categories of clothes each item is a 28 by 28 gray-scale pixel image  \cite{25}.
%	\item[4.] Glass : Contains 214 observations of glass shards. Each glass shard belongs to one of six classes which depends on 9 features that capture the chemical components of the glass shards these classes include float processed or non-float process building windows, vehicle windows, containers, table-ware, or head lamps   \cite{15}.
%	\item[5.] Cancer: Contains 699 observations each containing a tumor. Depending on the values of 30 features the observations are then classified into tumors either being benign or malignant \cite{15}. 
%\end{description}

The inputs values for all problems were standardised using the $Z$-score normalisation. All labels were binary encoded for each problem that contained two output classes, and one-hot encoded for problems that contained more than two output classes.

\subsection{ANN Hyper-parameters}\label{subsec:arch}

\begin{table}[!t]
	\caption{ANN architectures and hyper-parameter values used for each problem (Function: output layer activation function).}
	\label{tab:arch_param}	
\centering
\begin{tabular}{ |l|c|c|c|c|p{1.2cm}|p{1.2cm}|c|   }
	\hline
	Problem & Input & Hidden & Output & Dimension & Learning rate & Batch size & Function\\
	\hline
	Cancer~\cite{15} &30 &10 &1 &321& 0.0005    & 32 &   Sigmoid \\
	Glass~\cite{15}  &9 & 9&  6 & 150  &  0.005  & 32   & Softmax  \\
	Diabetes~\cite{15} &  8  & 8   &1 & 81 &0.0005  & 32 &  Sigmoid  \\
	MNIST~\cite{16} & 784    &10 &   10 & 7960 &0.001 & 128 &  Softmax \\
	Fashion-MNIST~\cite{25} &784 & 10&  10 & 7960 &0.0005 & 128 &  Softmax \\
	
	\hline
\end{tabular}
\end{table}

The ANN architecture that was adopted for each problem is given in Table~\ref{tab:arch_param}. The table summarises the ANN architecture used for each dataset, the source from which each dataset and/or ANN architectures was adopted, the total dimensionality of the weight space and the corresponding hyper-parameter sets. The hyper-parameters for each architecture was chosen based on the best performance for the given architecture in terms of accuracy and training time. The network size for each problem was chosen to be as minimal as possible based on the principle of parsimony, this was to ensure that the ANN captured the general characteristics of the datasets and that no ANN was over-parameterised.
%The ANN architecture that was adopted for each problem is given in Table~\ref{tab:arch_param} with the hyperparameter values. Each network size was chosen to be as minimal as possible, to ensure that the ANN captured the general characteristics of the datasets and that no ANN was over-parameterised.%

All experiments conducted used 10-fold cross validation, and each test was run 30 independent times for 100 epochs. The optimiser used for all experiments was Adam, as it is considered to be robust against suboptimal hyper-parameter choices, providing more leeway in parameter tuning~\cite{23}. The learning rates tested for Adam for each problem were $0.0005, 0.001$, and $0.005$. The best-performing learning rate was then chosen based on the generalisation performance across all loss functions for a problem. The activation function that was used for all experiments in the hidden layer was the Relu activation function, as it is considered the standard activation function to use in practice for neural networks at the time of writing \cite{27}. 

\section{Results}\label{sec:results}
\begin{figure}[!t]
\centering
\subfloat[Cancer]{\includegraphics[width=2.358in]{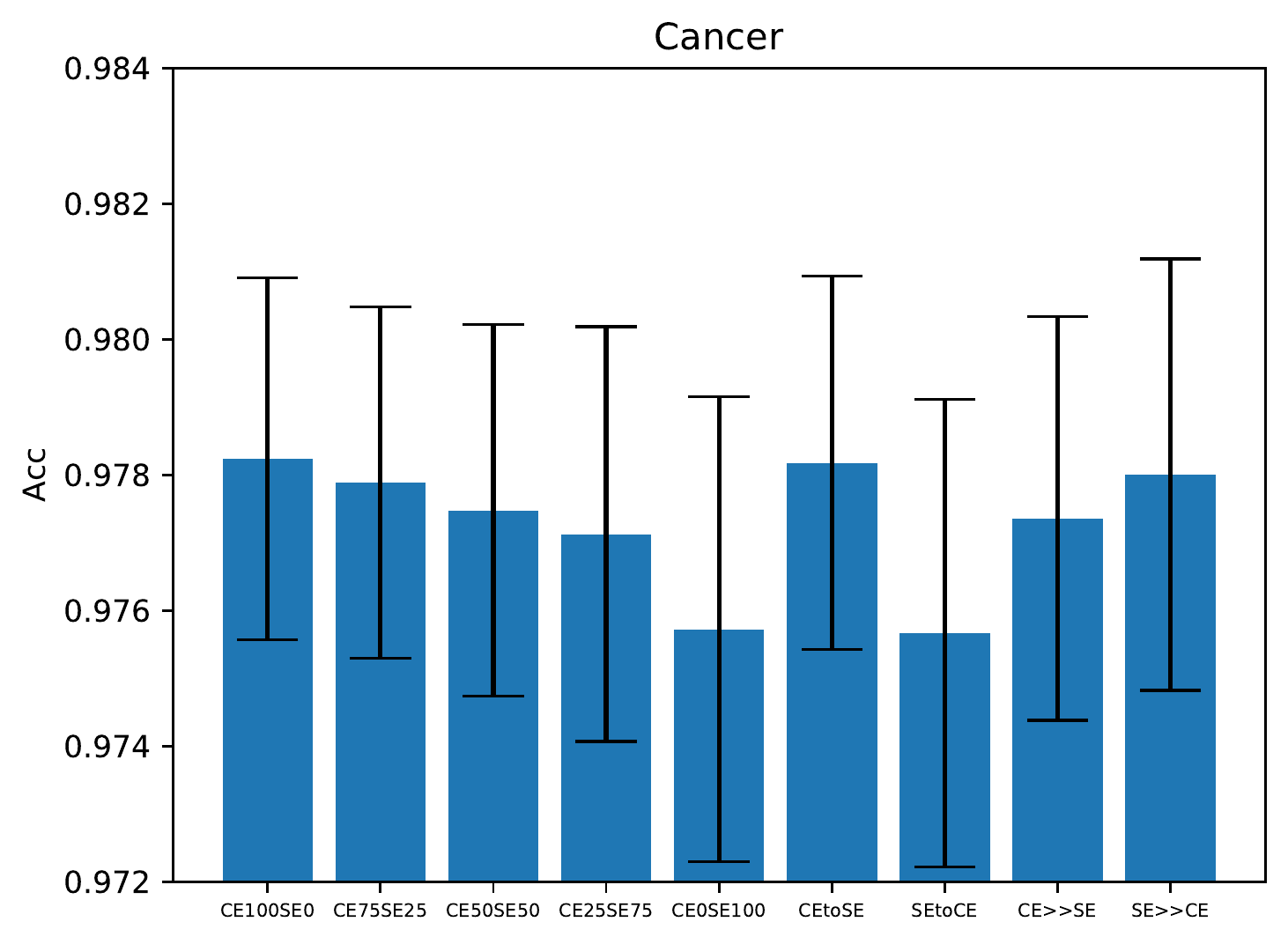}
\label{fig:bar_cancer}}
\hfil
\subfloat[Glass]{\includegraphics[width=2.358in]{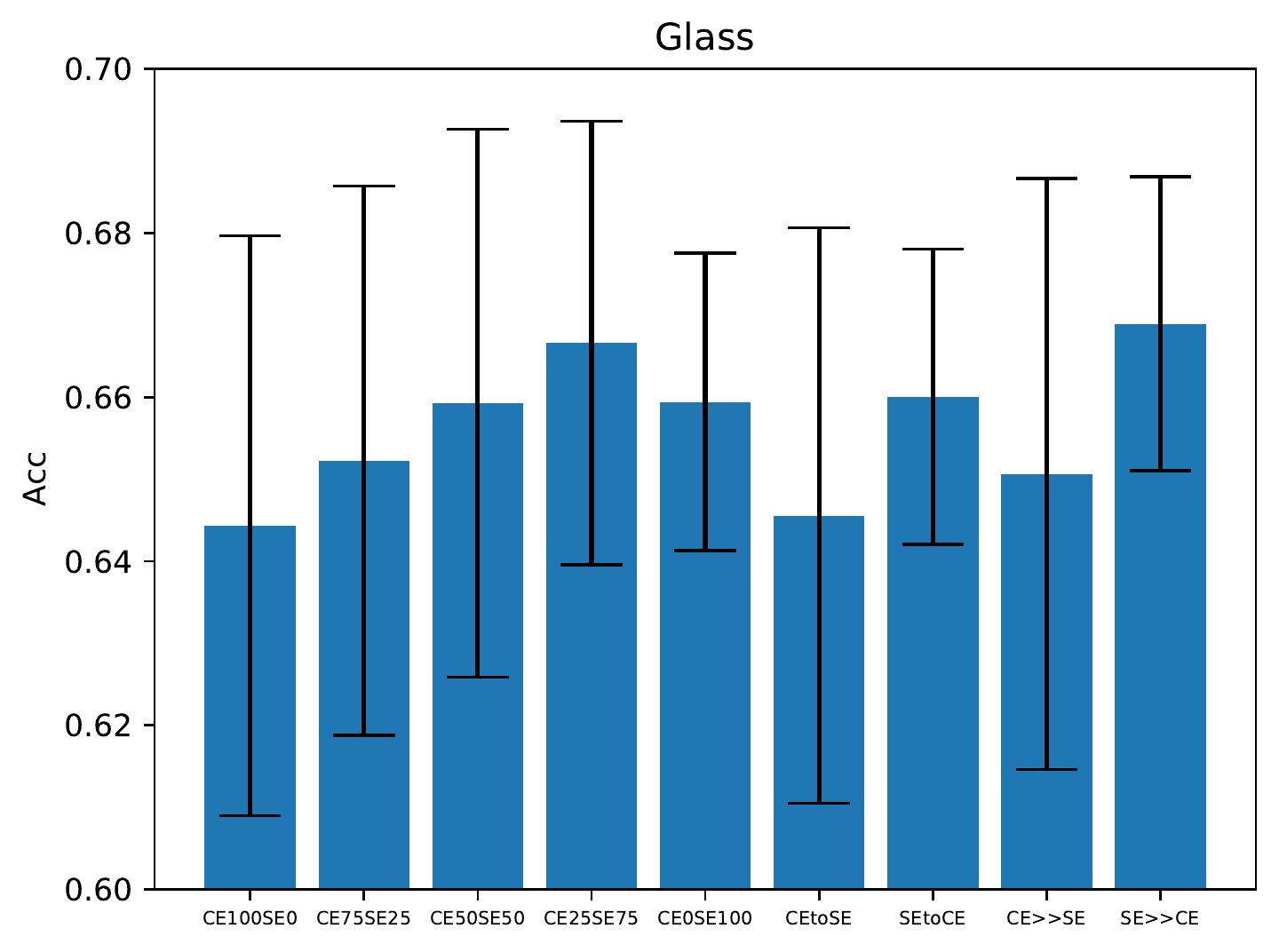}
\label{fig:bar_glass}}\\
\subfloat[Diabetes]{\includegraphics[width=2.358in]{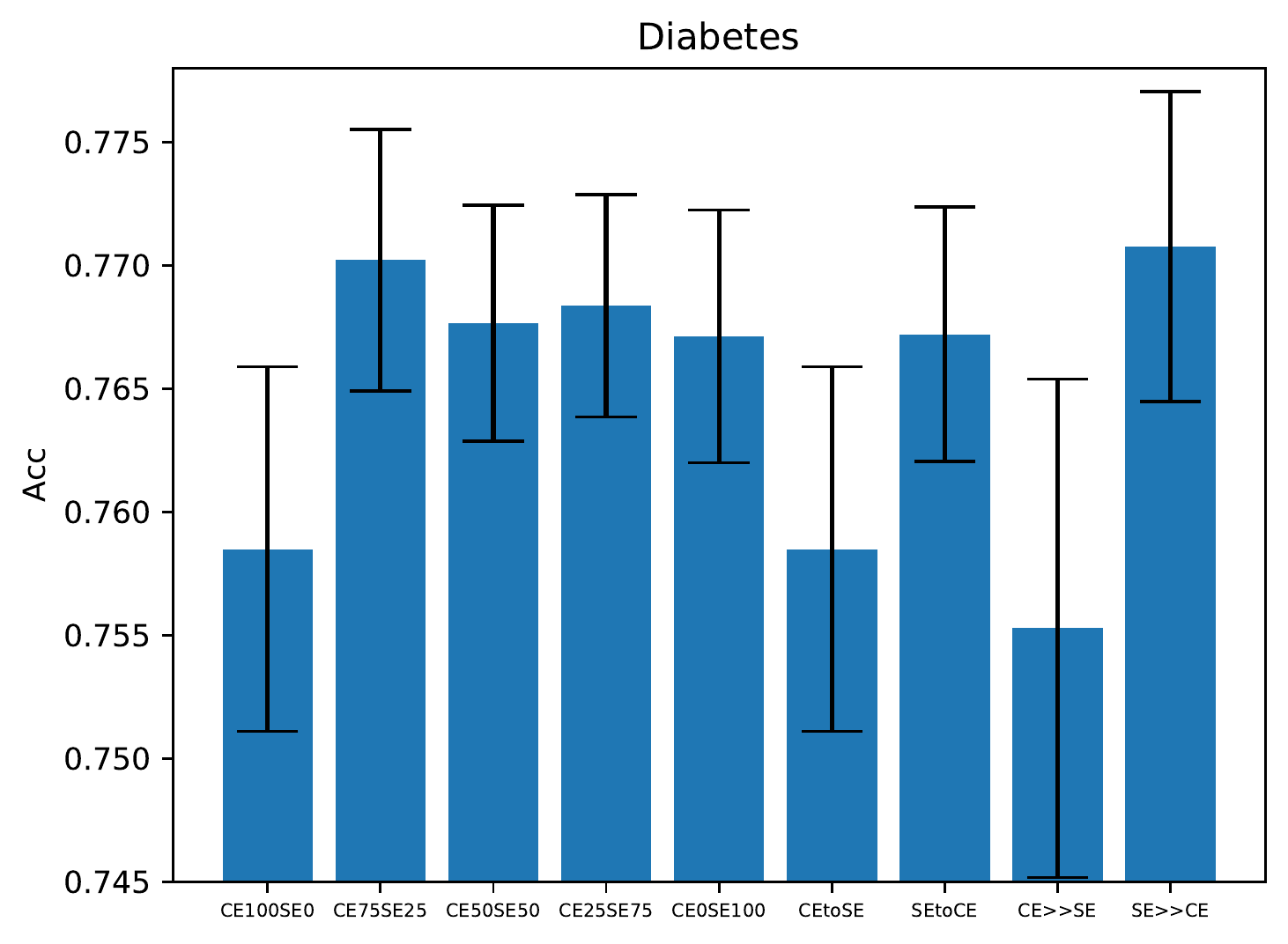}
\label{fig:bar_diabetes}}
\hfil
\subfloat[MNIST]{\includegraphics[width=2.358in]{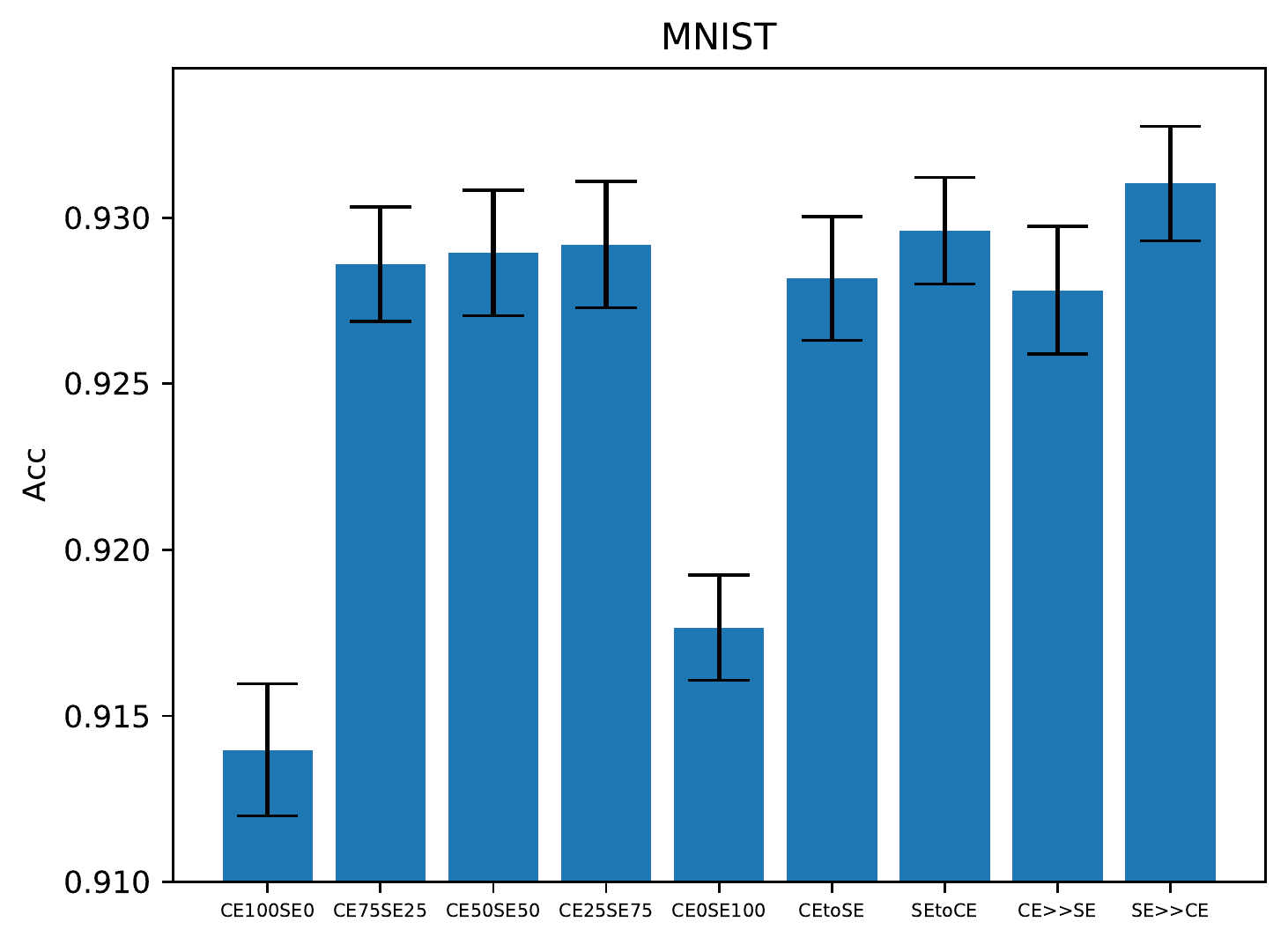}
\label{fig:bar_mnist}}\\
\subfloat[Fashion-MNIST]{\includegraphics[width=2.358in]{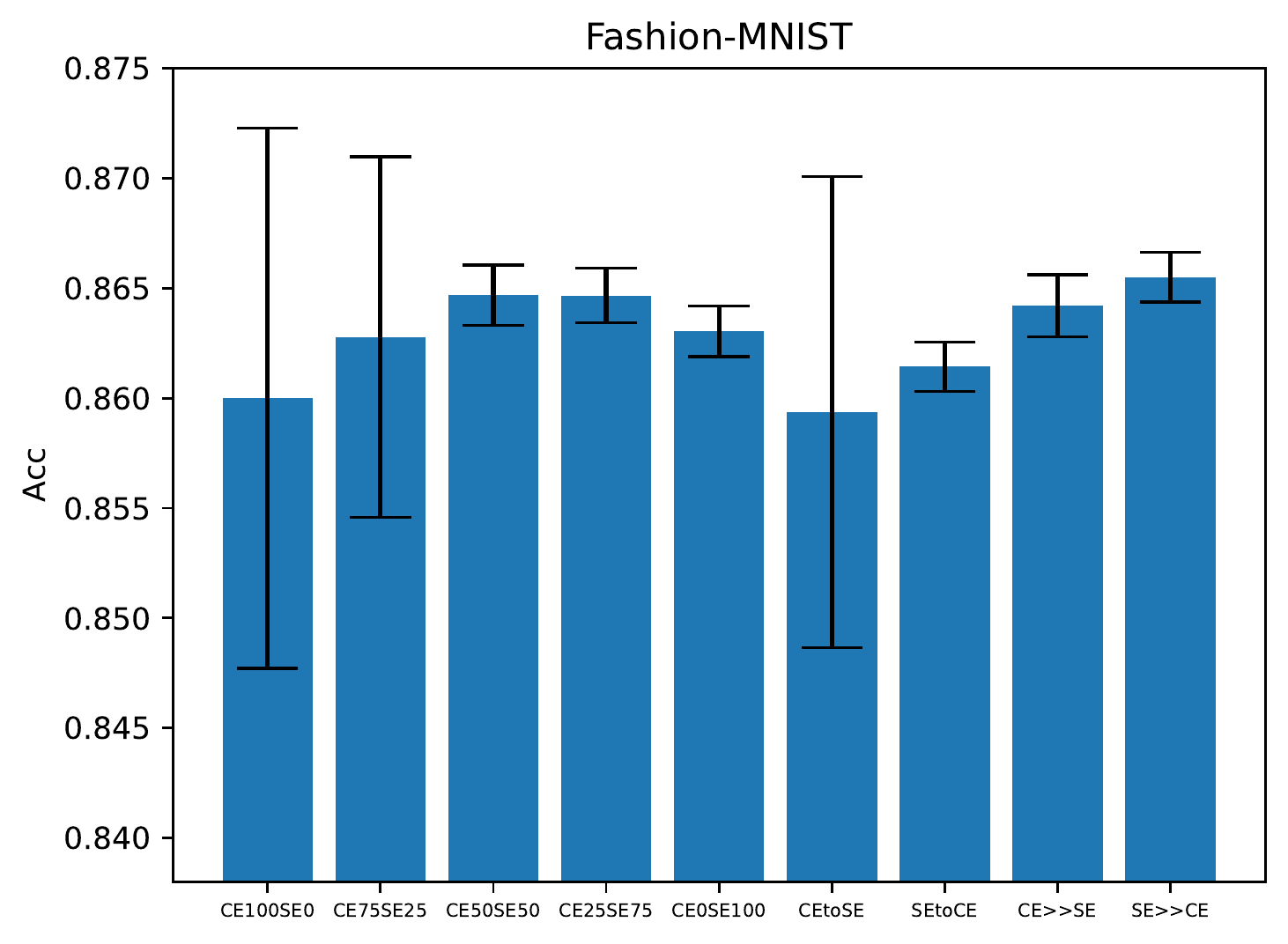}
\label{fig:bar_fmnist}}
\caption{Average accuracy with standard deviation bars on the test set for the five classification problems when using the nine loss function variants. Each bar represents a variant plotted in order from left to right (1) $CE_{100}SE_{0}$, (2) $CE_{75}SE_{25}$, (3) $CE_{50}SE_{50}$, (4) $CE_{25}SE_{75}$, (5) $CE_{0}SE_{100}$, (6) $CE_{to}SE$, (7) $SE_{to}CE$, (8) $CE_{>>}SE$, and (9) $SE_{>>}CE$.}
\label{fig:bars}
\end{figure}

Figure \ref{fig:bars} shows the average test accuracy with standard deviation bars on the five datasets. The bars in each plot represent the accuracy for the nine variants of loss functions in the same order as they appear in Table \ref{tab:lossfunctions}: (1) $CE_{100}SE_{0}$, (2) $CE_{75}SE_{25}$, (3) $CE_{50}SE_{50}$, (4) $CE_{25}SE_{75}$, (5) $CE_{0}SE_{100}$, (6) $CE_{to}SE$, (7) $SE_{to}CE$, (8) $CE_{>>}SE$, and (9) $SE_{>>}CE$. 

For example, Figure \ref{fig:bar_mnist} shows that the lowest accuracy values were achieved for the first and fifth variants which correspond to the CE only and SE only loss functions. This shows that in the case of the MNIST dataset, any of the hybrids performed better than either of the constituent loss functions on their own. Although the last variant resulted in slightly higher accuracy values than the other hybrids, there is not much difference in the performance. 

The standard deviation bars in Figure \ref{fig:bars} show that some variants resulted in higher deviations in performance than other variants. For example, Figure \ref{fig:bar_fmnist} shows that on the Fashion-MNIST dataset, the first (CE only), second (static hybrid with 75\% CE) and sixth (adaptive hybrid starting with 100\% CE) variants had wider deviations in accuracy values than the other variants. This indicates that the use of the CE loss function results in more volatile performance on the Fashion-MNIST dataset.

The mean test accuracy values are presented in Table \ref{tab:accuracy} with standard deviation values given in parentheses below each mean. The Mann-Whitney U test~\cite{mann1947test} was carried out in order to compare the average test accuracy results of all loss functions to one another to establish if the difference was statistically significant. The null hypothesis $H_0 : \mu_1 = \mu_2$, where $\mu_1$ and $\mu_2$
are the means of the two samples being compared, was evaluated at a significance level of 95\%. The alternative hypothesis was defined as $H_1 : \mu_1 \neq \mu_2$. Any p-value $< 0.05$ corresponded to rejection of the null hypothesis, and indicated statistical significance.

Mean values plotted in red indicate the worst performance on that dataset and values plotted in green indicate the best performing hybrid. Results that were not statistically significantly different shared the best or worst result. For example, on the Cancer dataset, all of the variants performed equally well except for the fifth (SE only) and seventh (adaptive hybrid starting with 100\% SE) variants, that performed equally poorly. This result is confirmed in Figure \ref{fig:bar_cancer}, where the fifth and seventh bars are clearly lower than the others, even though the standard deviations are very large across all variants. 
From the results in Table \ref{tab:accuracy}, it can be seen that for all datasets except Cancer, a loss function using only CE ($CE_{100}SE_{0}$) was the worst performing option. In addition, for two of the datasets, a loss function using only SE ($CE_{0}SE_{100}$) was the worst performing option. It can also be seen that for all datasets, $SE_{>>}CE$ (reactive hybrid that started with SE and switched to CE on stagnation/deterioration) was the best performing loss function variant.

\begin{table*}[!t]
\renewcommand{\arraystretch}{1}
\caption{Average test accuracy with corresponding standard deviations for all loss function variants and problems }
\label{tab:accuracy}\centering
\begin{tabular}{|c|c|c|c|c|c|} 
 \hline
  & Cancer & Glass & Diabetes & MNIST & Fashion-MNIST\\ \hline
\multirow{2}{*}{$CE_{100}SE_{0}$}  
  &  \color{ForestGreen}0.9782 & \color{red}0.6443 & \color{red}0.7585 & \color{red}0.9140 & \color{red}0.8600\\
  & ($\pm 0.0.0027$) & ($\pm 0.0353$) & ($\pm 0.0074$)& ($\pm 0.0020$) & ($\pm 0.0123$)\\ \hline  
  
\multirow{2}{*}{$CE_{75}SE_{25}$}  
  & \color{ForestGreen}0.9779 & 0.6523 & \color{ForestGreen}0.7702 & 0.9286 & 0.8628\\ 
  & ($\pm 0.0026$) & ($\pm 0.0335$) & ($\pm 0.0053$) & ($\pm 0.0017$) & ($\pm 0.0082$)\\ \hline
  
\multirow{2}{*}{$CE_{50}SE_{50}$} 
  &  \color{ForestGreen}0.9775 & \color{ForestGreen}0.6593 & 0.7677 & 0.9289& 0.8647\\ 
  &  ($\pm 0.0027$) & ($\pm 0.0334$) & ($\pm 0.0048$) & ($\pm 0.0019$)& ($\pm 0.0014$)\\ \hline
  
\multirow{2}{*}{$CE_{25}SE_{75}$}  
  &  \color{ForestGreen}0.9771 & \color{ForestGreen}0.6666 & 0.7684 & 0.09292& 0.8647\\ 
  & ($\pm 0.0031$) & ($\pm 0.0270$) & ($\pm 0.0045$) & ($\pm 0.0019$) & ($\pm 0.0012$) \\ \hline

\multirow{2}{*}{$CE_{0}SE_{100}$} 
  & \color{red}0.9757 & 0.6594 & 0.7671 & 0.9177 & \color{red}0.8630\\ 
  & ($\pm 0.0034$) & ($\pm 0.0181$) & ($\pm 0.0051$)& ($\pm 0.0016$) & ($\pm 0.0012$)\\ \hline
  
\multirow{2}{*}{$CE_{to}SE$}
  & \color{ForestGreen}0.9782 & 0.6456 & \color{red}0.7585 & 0.9282 & \color{red}0.8594\\ 
  & ($\pm 0.0028$) & ($\pm 0.0351$) & ($\pm 0.0074$) & ($\pm 0.0019$)& ($\pm 0.0107$)\\ \hline
  
\multirow{2}{*}{$SE_{to}CE$}
  & \color{red}0.9757 & 0.6601 & 0.7672 & 0.9296 & 0.8614\\
  & ($\pm 0.0035$) & ($\pm 0.0180$)& ($\pm 0.0052$)& ($\pm 0.0016$)& ($\pm 0.0011$)\\ \hline
  
\multirow{2}{*}{$CE_{>>}SE$}
  & \color{ForestGreen}0.9774& 0.6506 & \color{red}0.7553 & 0.9278 & 0.8642\\
  & ($\pm 0.0030$) & $(\pm 0.0360$) & ($\pm 0.00101$)& ($\pm 0.0019$) & ($\pm 0.0014$)\\ \hline
  
\multirow{2}{*}{$SE_{>>}CE$}
  &\color{ForestGreen}0.9780 & \color{ForestGreen}0.6689 & \color{ForestGreen}0.7708 & \color{ForestGreen}0.9310 & \color{ForestGreen}0.8655\\ 
  & ($\pm 0.0032$) & ($\pm 0.0179$) & ($\pm 0.0063$) & ($\pm 0.0017$) & ($\pm 0.0011$)\\ \hline
\end{tabular}
\end{table*}

\section{Discussion}\label{sec:discuss}
Results showed that on the datasets studied, hybrid loss functions generalise better than the baseline loss functions CE and SE. This provides evidence to the claim made by Bosman \emph{et al.} \cite{2} that the SE and CE loss functions should be combined in one way or another, as their error landscapes were shown to exhibit different yet complementary properties. In~\cite{2}, the CE loss function was shown to be more searchable, and the SE loss function was shown to be more robust against overfitting.

%Of all the hybrid functions, the best performing variant overall was the $CE_{>>}SE$ function (either the best performing or not statistically significantly different to the best performing variant). This variant starts with the CE function and switches to SE once stagnation or deterioration in accuracy is detected. This confirms results by Golik \emph{et al.} \cite{3} who found that SE did not perform well with randomised weights, but performed well when used to fine tune the solution that the CE loss function produced.

Of all the hybrid functions, the best performing variant overall was the $SE_{>>}CE$ function (either the best performing or not statistically significantly different from the best performing variant). This variant starts with the SE function, and switches to CE once stagnation or deterioration in accuracy is detected. This is contrary to work done by Golik \emph{et al.} \cite{3} who analysed that starting with CE loss function and later switching over to the SE loss function produced superior generalised models than SE and CE loss functions individually. This was due to the SE loss function not performing well with randomised weights, but performed well when used to fine tune the solution that the CE loss function produced, this study confirms the results found by Golik. However, during experimentation it was shown that the minima discovered by SE loss function can be further exploited by switching to CE loss function. Golik found that tuning hyper-parameters such as the learning rate and batch size had some effect on switching the CE loss function to the SE loss function while other hyper-parameters were not very sensitive.

%Therefore, beginning with the SE loss function allows a gradient based algorithm to be less likely to oscillate between the walls of such valleys which could contain an area of poor fitness. Instead it guides the gradient descent algorithm into areas of good fitness which are found lower down the valley of the loss landscape, less able to be reached by the SE loss function.

The landscapes of the CE loss function have been shown to contain narrower valleys, but fewer local optima than SE loss landscapes \cite{2}. Thus, a gradient-based algorithm starting at an initial random position is more likely to descend into a narrow valley during training, which is associated with poor generalisation performance~\cite{chaudhari2019entropy,keskar2016large}. On the other hand, SE has been shown to be more resilient to overfitting than CE \cite{2}. Therefore, starting with SE, the search algorithm has a better chance of reaching a better region in terms of generalisation. Switching then to CE, the stronger gradients can assist the algorithm to exploit the discovered minimum. 

In other words, during training, the SE loss function focuses on exploring an area in the fitness landscape until it finds itself stuck at some arbitrary point. Once it is stuck the switch to the CE loss function occurs where the CE loss function then exploits the solution obtained by the SE loss function. This process seems to exhibit a generalisation ability that is superior to the base line loss functions and all other hybrid loss functions tested.

\begin{figure}[!t]
\centering
\subfloat[Cancer]{\includegraphics[width=2.28in]{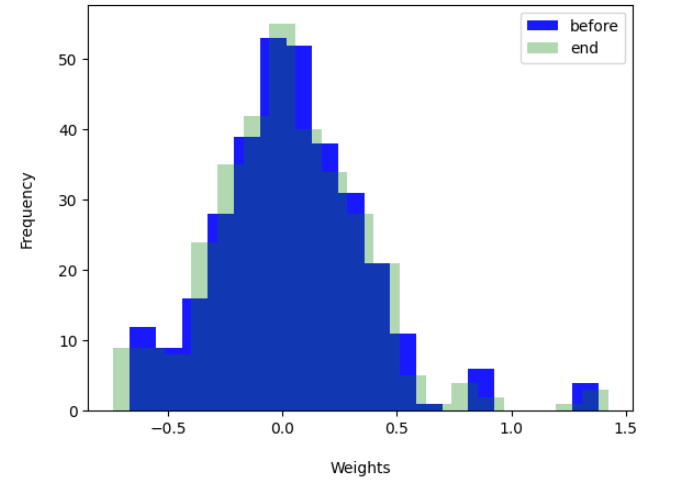}
\label{fig:hist_cancer}}
\hfil
\subfloat[Glass]{\includegraphics[width=2.28in]{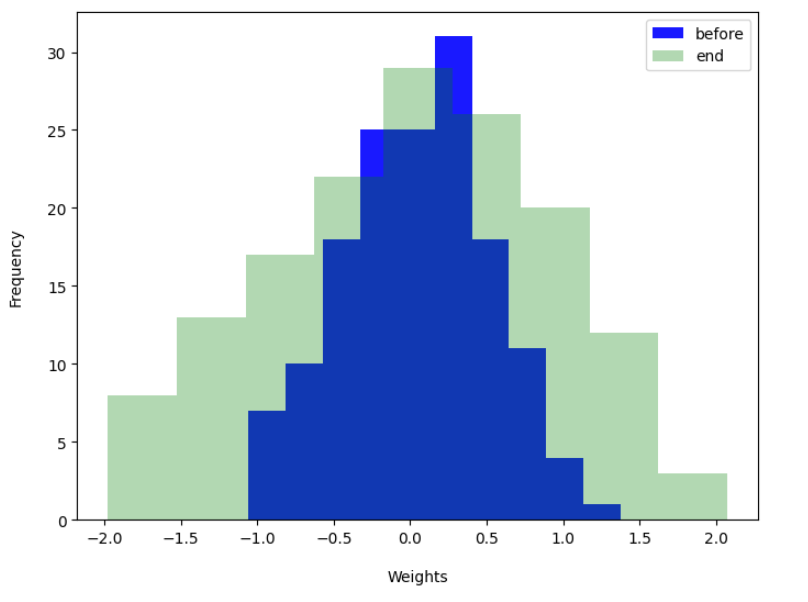}
\label{fig:hist_glass}}\\
\subfloat[Diabetes]{\includegraphics[width=2.28in]{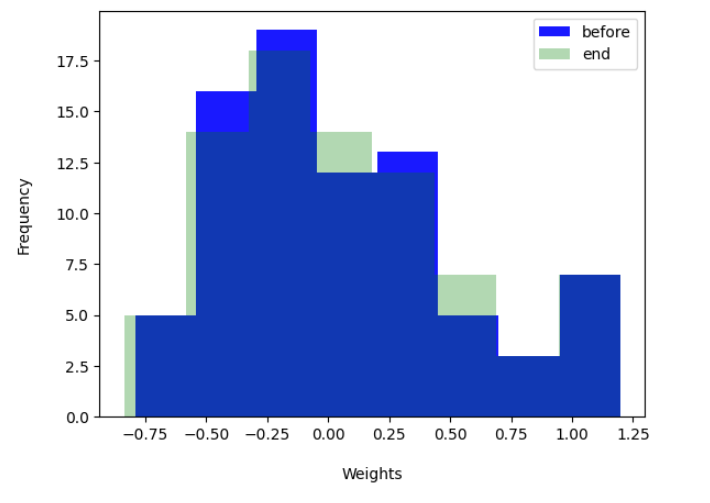}
\label{fig:hist_diabetes}}
\hfil
\subfloat[MNIST]{\includegraphics[width=2.28in]{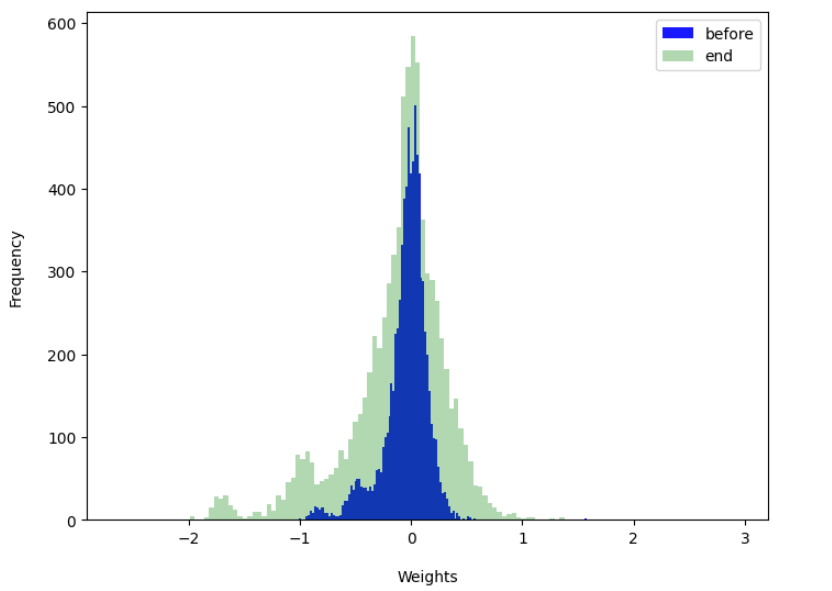}
\label{fig:hist_mnist}}\\
\subfloat[Fashion-MNIST]{\includegraphics[width=2.28in]{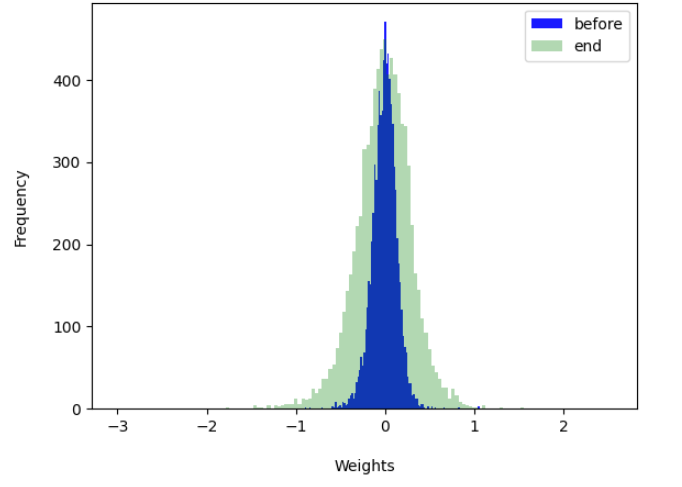}
\label{fig:hist_fmnist}}
\caption{Histograms showing the distribution of neural network weights for the $SE_{>>}CE$ loss function variant before the switch (blue) and after training (light green).}
\label{fig:hist}
\end{figure}

To further analyse the behaviour of the $SE_{>>}CE$ variant, Figure~\ref{fig:hist} shows the distribution of neural network weights before the switch (blue) and at the end of training (light green) for the five problems. For example, Figure \ref{fig:hist_cancer} shows that before the switch (when SE was being utilised as the loss function), most of the weights had values around zero, but there were a few weights with values above 1.    

The histograms in Figure~\ref{fig:hist} show that the weights exhibited a negative skew, indicating that a larger proportion of weights were set to negative values at the end of training. The hidden layer activation function used in this study was the Relu function, which outputs a zero for negative input. Thus, a large number of negative weights indicates that more neurons were effectively switched off (eg. output set to zero) as the training progressed. Disabling unnecessary neurons can be seen as a form of self-regularisation exhibited by the ANN during training. This self-regularising behaviour may be responsible for the superior generalisation performance of the $SE_{>>}CE$ hybrid.

\section{Conclusion}\label{sec:conclude}
This paper proposed to hybridise the squared loss function (SE) with the entropic loss function (CE) to create a new loss function with superior properties. Three types of hybrids were proposed: static, adaptive, and reactive. Experiments were conducted to determine the effectiveness of the proposed hybrids.

The hybridisation of the SE and CE loss functions was shown to improve the generalisation ability for all five problems considered compared to the performance exhibited by the baseline loss function CE and SE. It was also concluded that the hybrid loss function $SE_{>>}CE$, which started with the SE loss function and then switched over to the CE loss function, performed the best on average or was not significantly different from the best for all problems tested. It was also observed that the hybrid loss functions that started with or placed more importance on the CE loss function exhibited higher standard deviation throughout training, showing CE to be more volatile in regards to the initialisation of weights. The opposite configuration of baseline loss functions was shown to be more robust against the initialisation of weights in the model. Therefore, the robustness of the hybrid loss functions that start with or give more importance to the SE loss function should be considered when hybridising CE and SE in order to produce optimal hybrid loss functions.

Future work will include a scalability study of the proposed hybrids. Performance of $SE_{>>}CE$ can be evaluated on various modern deep ANN architectures. Fitness landscape analysis techniques can be employed to study the landscapes of the proposed hybrids.

\newpage
\bibliographystyle{splncs04}
%\bibliography{M335}

\begin{thebibliography}{10}
\providecommand{\url}[1]{\texttt{#1}}
\providecommand{\urlprefix}{URL }
\providecommand{\doi}[1]{https://doi.org/#1}

\bibitem{2}
Bosman, A.S., Engelbrecht, A., Helbig, M.: Visualising basins of attraction for
  the cross-entropy and the squared error neural network loss functions.
  Neurocomputing  \textbf{400},  113--136 (2020).
  \doi{10.1016/j.neucom.2020.02.113}

\bibitem{9}
Bourlard, H.A., Morgan, N.: Connectionist Speech Recognition. Springer {US}
  (1994). \doi{10.1007/978-1-4615-3210-1},
  \url{https://doi.org/10.1007/978-1-4615-3210-1}

\bibitem{chaudhari2019entropy}
Chaudhari, P., Choromanska, A., Soatto, S., LeCun, Y., Baldassi, C., Borgs, C.,
  Chayes, J., Sagun, L., Zecchina, R.: Entropy-{SGD}: Biasing gradient descent
  into wide valleys. Journal of Statistical Mechanics: Theory and Experiment
  \textbf{2019}(12),  124018 (2019)

\bibitem{3}
Golik, P., Doetsch, P., Ney, H.: Cross-entropy vs. squared error training: a
  theoretical and experimental comparison. In: 14th Annual Conference of the
  International Speech Communication Association. pp. 1756--1760. ISCA (2013)

\bibitem{keskar2016large}
Keskar, N.S., Mudigere, D., Nocedal, J., Smelyanskiy, M., Tang, P.T.P.: On
  large-batch training for deep learning: Generalization gap and sharp minima.
  arXiv preprint arXiv:1609.04836  (2016)

\bibitem{23}
Kingma, D., Ba, J.: Adam: A method for stochastic optimization. In:
  International Conference on Learning Representations (2015)

\bibitem{8}
Li, H., Xu, Z., Taylor, G., Studer, C., Goldstein, T.: Visualizing the loss
  landscape of neural nets. In: Proceedings of the 32nd Conference on Neural
  Information Processing Systems. pp. pages 6391--6401 (2018)

\bibitem{mann1947test}
Mann, H.B., Whitney, D.R.: On a test of whether one of two random variables is
  stochastically larger than the other. The annals of mathematical statistics
  pp. 50--60 (1947)

\bibitem{27}
Nwankpa, C., Ijomah, W., Gachagan, A., Marshall, S.: Activation functions:
  Comparison of trends in practice and research for deep learning. ArXiv
  \textbf{abs/1811.03378} (2018)

\bibitem{15}
Prechelt, L.: {PROBEN1} - {A} {Set} {of} {Neural} {Network} {Benchmark}
  {Problems} {and} {Benchmarking} {Rules}. Tech. Rep. Technical Report 21/94
  (07 1995)

\bibitem{22}
Solla, S.A., Levin, E., Fleisher, M.: Accelerated learning in layered neural
  networks. Complex Systems  \textbf{2},  pages 625--640 (1988)

\bibitem{25}
Xiao, H., Rasul, K., Vollgraf, R.: Fashion-mnist: a novel image dataset for
  benchmarking machine learning algorithms. ArXiv  \textbf{abs/1708.07747}
  (2017)

\bibitem{16}
Yann~LeCun, Corinna~Cortes, C.B.: Mnist handwritten digit database,
  \url{http://yann.lecun.com/exdb/mnist/}

\end{thebibliography}

%
\end{document}